\documentclass{article}

\usepackage{arxiv}
\usepackage{float} % for 'H' float specifier
\usepackage[utf8]{inputenc} % allow utf-8 input
\usepackage[T1]{fontenc}    % use 8-bit T1 fonts
\usepackage{url}            % simple URL typesetting
\usepackage{booktabs}       % professional-quality tables
\usepackage{amsfonts}       % blackboard math symbols
\usepackage{nicefrac}       % compact symbols for 1/2, etc.
\usepackage{microtype}      % microtypography
\usepackage{lipsum}		% Can be removed after putting your text content
\usepackage{graphicx}
\usepackage{natbib}
\usepackage{doi}
\usepackage{tabularx} % Add this in the preamble of your document
\usepackage{booktabs} 
\usepackage{algorithm}
\usepackage{algpseudocode}
\usepackage{amsmath}
\usepackage{placeins}
\usepackage{makecell} % Include this package to use the makecell command

\title{KARINA: An Efficient Deep Learning Model for Global Weather Forecast}

\author{Minjong Cheon \and Yo-Hwan Choi \and Seon-Yu Kang \and Yumi Choi \and Jeong-Gil Lee \and Daehyun Kang\thanks{Corresponding author. Email: \texttt{dkang@kist.re.kr}}}

\date{Center for Sustainable Environment Research\\ Korea Institute of Science and Technology\\ Seoul, South Korea}

\renewcommand{\shorttitle}{\textit{arXiv} Template}

\hypersetup{
pdftitle={A template for the arxiv style},
pdfsubject={q-bio.NC, q-bio.QM},
pdfauthor={Minjong},
pdfkeywords={First keyword, Second keyword, More},
}

\begin{document}
\maketitle

\begin{abstract}
Deep learning-based, data-driven models are gaining prevalence in climate research, particularly for global weather prediction. However, training the global weather data at high resolution requires massive computational resources. Therefore, we present a new model named KARINA to overcome the substantial computational demands typical of this field. This model achieves forecasting accuracy comparable to higher-resolution counterparts with significantly less computational resources, requiring only 4 NVIDIA A100 GPUs and less than 12 hours of training. KARINA combines ConvNext, SENet, and Geocyclic Padding to enhance weather forecasting at a 2.5° resolution, which could filter out high-frequency noise. Geocyclic Padding preserves pixels at the lateral boundary of the input image, thereby maintaining atmospheric flow continuity in the spherical Earth. SENet dynamically improves feature response, advancing atmospheric process modeling, particularly in the vertical column process as numerous channels. In this vein, KARINA sets new benchmarks in weather forecasting accuracy, surpassing existing models like the ECMWF S2S reforecasts at a lead time of up to 7 days. Remarkably, KARINA achieved competitive performance even when compared to the recently developed models (Pangu-Weather, GraphCast, ClimaX, and FourCastNet) trained with high-resolution data having 100 times larger pixels. Conclusively, KARINA significantly advances global weather forecasting by efficiently modeling Earth's atmosphere with improved accuracy and resource efficiency.
\end{abstract}

\section{Introduction}
\label{sec:intro}

Numerical Weather Prediction (NWP) has been a cornerstone of weather forecasting over the past decades due to substantial improvements in physics parameterization methods and the incorporation of high-quality observations and data assimilation \citep{gustafsson2018survey,bannister2017review}. Despite these developments, the significant processing power needed for NWP models has led to the development of data-driven approaches that make use of deep learning and machine learning as feasible replacements \citep{ren2021deep, weyn2021sub}. While these innovative approaches yielded significant advancements, their adoption was constrained by the high computational demands necessary for training. This could be illustrated by examples such as Pangu Weather, which required 192 Nvidia V100 GPUs for a training period of 64 days, and Fengwu, which utilized 32 Nvidia A100 GPUs over 17 days, underscoring the extensive resources needed for the development of cutting-edge, data-driven weather forecasting models \citep{bi2023accurate,chen2023fengwu}.

In this study, we introduce a model named KIST’s Atmospheric Rhythm with Integrated Neural Algorithms (KARINA), a global weather forecasting model developed at a 2.5° resolution that combines ConvNext with Squeeze-and-Excitation Networks (SENet) technologies to enhance prediction accuracy. Drawing inspiration from the FuXi model's success with the Swin transformer, we chose it as our foundation, incorporating ConvNext for its compatible design features \citep{chen2023fuxi}. We innovatively integrate SENet within ConvNext's architecture for channel recalibration and introduce a novel concept termed the `Geocyclic Padding' technique to improve spatial data handling \citep{iandola2016squeezenet}. This approach improves the model's predictive performance by prioritizing informative variables and maintains efficiency, making it an ideal enhancement for handling the complexity of high-dimensional climate data. Our experiments show that a ConvNext-based model, inherently biased towards processing high-dimensional spatial data efficiently, excels with the ERA5 daily dataset \citep{aniraj2023masking}. 

\begin{itemize}
    \setlength\itemsep{0.5em} % Adjust 1em to the desired space

    \item We present KARINA, an advanced iteration of the ConvNext architecture, which demonstrates superior performance over subseasonal-to-seasonal (S2S) reforecasts from the European Centre for Medium-Range Weather Forecasts (ECMWF) Integrated Forecasting System (IFS) and achieves parity with other state-of-the-art (SOTA) models within the 7-day prediction.
    
    \item Our proposed Geocyclic Padding and SENet capture the salient features of atmospheric physics on the spherical Earth, resulting in marked improvements at the image edges and the equatorial regions.
    
    \item KARINA sets itself apart from other SOTA models for computational efficiency, delivering superior global weather predictions with under 12 hours of training on 4 NVIDIA A100 GPUs.
\end{itemize}

\section{Related Works}

\subsection{CNN-based models}
At first, ResNet, which is an algorithm based on Convolution Neural Network (CNN) was applied to predict several atmospheric variables, including 500 hPa geopotential (Z500), and 850 Pa temperature (T850) up for 5 days \citep{rasp2020weatherbench}. Weyn et al. introduced an advanced global weather forecasting model using deep convolutional neural networks, achieving significant improvements by employing a cubed-sphere grid and enhanced CNN architecture for stable and realistic forecasts for several weeks \citep{weyn2020improving}. Rasp and Thuerey explored medium-range weather forecasting through a deep residual convolutional neural network, focusing on predicting geopotential, temperature, and precipitation \citep{rasp2021data}. 

\subsection{VIT-based models}
FourCastNet leveraged the Adaptive Fourier Neural Operator networks for its forecasting tasks and implemented a dual-phase finetuning process to enhance the precision of its autoregressive multi-step forecasting \citep{pathak2022fourcastnet}. Lam et al. developed "GraphCast," a high-resolution, graph neural network model for 10-day global weather forecasting. It effectively surpassed conventional models in predicting hundreds of variables with accuracy, particularly in forecasting severe weather conditions \citep{lam2022graphcast}. ClimaX also utilized the transformer architecture with novel encoding and aggregation blocks and this model finally enables effective pretraining on diverse datasets and generalizing well to a variety of downstream tasks in climate science \citep{nguyen2023climax}. Pangu-Weather employed the 3D Swin-Transformer architecture and introduced a method of hierarchical temporal aggregation to execute a variety of downstream forecasting scenarios, such as forecasts for extreme weather \citep{bi2023accurate}. Meanwhile, FengWu approached the forecasting challenge through a multi-task optimization lens, unveiling an innovative fine-tuning methodology known as the replay buffer to conduct global medium-range weather forecasts efficiently \citep{chen2023fengwu}. The SwinVRNN, a Swin Transformer-based Variational Recurrent Neural Network, enhanced medium-range weather forecasting accuracy and ensemble spread by combining deterministic predictions with stochastic perturbations, outperforming traditional models and operational standards like the ECMWF IFS for certain variables up to 5 days ahead \citep{hu2023swinvrnn}. The SwinRDM model, integrating an improved SwinRNN with a diffusion model, achieved high-resolution weather forecasting at 0.25° resolution, surpassing the accuracy of the SOTA IFS model for key atmospheric variables \citep{chen2023swinrdm}. Lastly, FuXi aimed to diminish the cumulative forecasting error across multiple steps by integrating a series of three U-Transformer models (short, medium, long) based on the Swin Transformer, each fine-tuned for optimal performance at distinct forecasting intervals \citep{chen2023fuxi}.

Leveraging the foundational advancements in neural network architectures for weather forecasting, which span from convolutional neural networks to graph neural networks and transformers, our study introduces KARINA based on the ConvNext architecture. In addition, we present coherent approaches to enhance the accuracy of weather forecasting. KARINA incorporates Geocyclic Padding to preserve the integrity of atmospheric flows, consistent with Earth's spherical shape. Simultaneously, SENet helps make dynamic feature adjustment possible, significantly improving the atmospheric column processes in the vertical pressure levels. This study first shows region-dependent skill improvement and variable relationships from each methodology.

\section{Methodology}

\subsection{Data Description}
The ECMWF Reanalysis v5 (ERA5), a reanalysis that covers a global atmosphere at 0.25° horizontal resolution and 37 vertical levels, was used as a dataset in this research \citep{hersbach2020era5}. We selected a total of 66 variables consisting of six surface variables and five variables with 12 vertical pressure levels from the surface to the lower stratosphere, known as prognostic variables of the Earth's atmosphere. With 66 variables acquired in hourly intervals, an interpolated dataset to daily average at 2.5° horizontal resolution is organized as a tensor with dimensions (72 × 144 × 66). We noted that the coarser temporal and spatial resolutions help exclude high-frequency noise and reduce uncertainty in data accuracy, thereby enabling training on more salient features crucial for global weather prediction if high-resolution data is not necessary for the large-scale features \citep{kim2022added}. A previous experiment proved that employing the 66 variables besides orography is far more successful than FourCastNet's 20 variables \citep{cheon2024advancing}. For training and evaluation, the dataset spanning from 1979 to 2015 serves as the training set, with the years 2016 and 2017 designated as the validation set, and the year 2018 forming the test set. 

In comparison to the KARINA, we utilized the ECMWF S2S reforecast, which is known as the most skillful S2S modeling system using numerical solutions. We used a control simulation of the ECMWF IFS model version CY47R3 archived in the S2S database \citep{vitart2017subseasonal}. This S2S model produces 46-day forecasts twice a week (every Monday and Thursday) with 1.5° degree horizontal resolution. For the skill evaluation, daily-averaged variables were obtained in the database and then interpolated to 2.5° degree.

\subsection{KARINA Description}
In our study, we refined the architecture of ConvNext to develop KARINA, an iteration that promotes ConvNext's performance for weather forecasting at a 2.5° resolution. Our model incorporated two key innovations tailored for the ERA5 dataset: the integration of SENet for dynamic feature recalibration and the introduction of Geocyclic Padding to preserve geographical continuity across the 0° and 360° meridians and the poles. SENet improved the model's focus on relevant weather variables, while Geocyclic Padding addresses the challenge of representing the Earth's spherical geometry, ensuring the integrity of horizontal atmospheric fluxes for global weather predictions.

\subsection{Baseline: Modified ConvNext}
ConvNext stands as a significant evolution in convolutional neural network algorithms, drawing inspiration from the transformative success of the Swin Transformer architecture \citep{liu2022convnet}. Swin Transformer, known for its efficiency and scalability in processing image data through a hierarchical transformer model, has set new benchmarks in computer vision \citep{liu2021swin}. By integrating aspects of the Swin Transformer, ConvNext enhanced its convolutional framework to better handle the complexities of spatial hierarchies and dependencies in data. This fusion allows ConvNext to benefit from the adaptability and context-aware processing capabilities of transformers while maintaining the efficiency and locality advantages of CNNs.

Stem Layer: The architecture begins with a Stem layer, a starting point for feature extraction. This layer is a 2D convolutional layer with kernel size 3, stride 1, and padding 1, a configuration that crucially maintains the spatial dimensions of the input. Post the convolutional operation in the stem layer, a layer normalization is applied to stabilize the network by normalizing the features emerging from the convolution, preparing them for subsequent layers \citep{ba2016layer}.

Depth Scaling Layer: Following the Stem layer, our architecture features a series of depth scaling layers, organized into three stages within a loop that iterates over four distinct ranges. Unlike the conventional downsampling layers used in the original ConvNext model, which reduces spatial dimensions, our approach focuses on increasing channel depth while preserving the original spatial dimensions of the input data \citep{liu2022convnet}. LayerNorm is applied sequentially to the channels at the beginning of each depth scaling stage, with channel sizes increasing from 96 to 192, to 384, and finally to 768. This is followed by convolutional layers that progressively enlarge the channel dimensions from 96, moving through 192 and 384, to reach 768. Each layer in our network is configured with a kernel size of 3, a stride of 1, and padding that preserves the original input size, ensuring that spatial dimensions are consistently maintained. 

Block Layer: After the depth scaling layer, our architecture introduces a depthwise convolutional layer with a kernel size of 7 and padding width of 3, utilizing our Geocyclic Padding technique. This layer processes each input channel independently to improve spatial feature extraction efficiency. Subsequently, a Squeeze-and-Excitation (SE) layer, accompanied by layer normalization, is applied to refine channel-wise features by selectively emphasizing important channels and diminishing less relevant ones. Following layer normalization, the architecture employs two pointwise convolution layers: the first expands the channel dimensions by a factor of four. In contrast, the second restores them to their original size, effectively transforming channel-wise feature representations. The inclusion of the Gaussian Error Linear Unit (GELU) activation function is added \citep{hendrycks2016gaussian}, and a learnable scaling parameter is also utilized to adjust the output based on the feature's significance. The model also integrates residual connections and drop path regularization to facilitate information flow and mitigate overfitting, respectively \citep{hayou2021regularization,he2016deep}. The `depths' array further details the structure's complexity, which specifies the number of Block instances in each stage, exemplified by [3, 3, 9, 3] for a four-stage design.

The final stages include a first convolutional layer for further feature processing, maintaining the channel dimensions, and a second convolutional layer, which acts as the prediction head and maps the features to the same feature shape as the input dataset, effectively preparing the network for prediction.

\begin{figure}[h!]
	\centering
	\includegraphics[width=\textwidth]{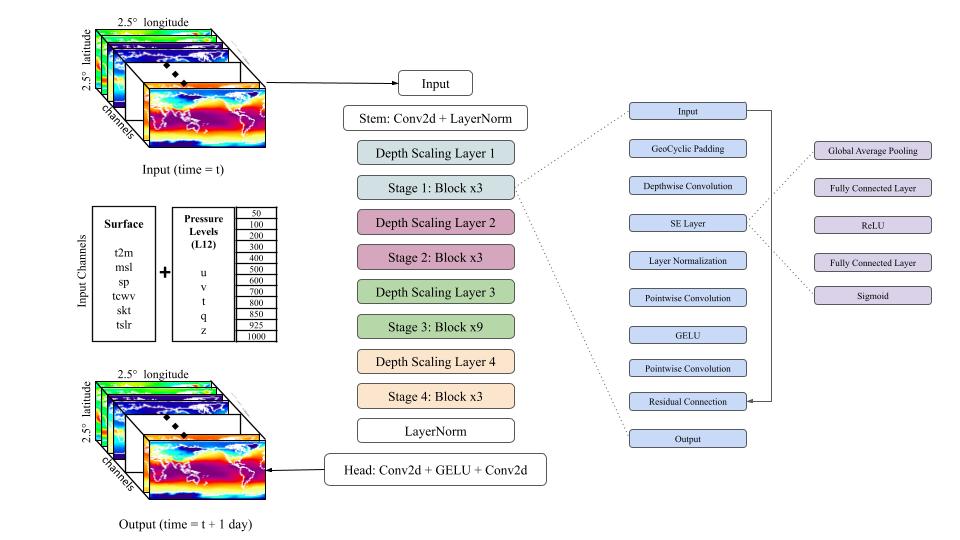}
	\caption{Overall description of KARINA, which merges ConvNext with SENet and Geocyclic Padding, targeting 2.5° forecasting precision. Features a [3, 3, 9, 3] stage design, with initial blocks using a kernel size of 7 and depth scaling layers employing a kernel size of 3, channel expansion from 96 to 768, preserving spatial consistency.}
	\label{fig:fig1}
\end{figure}

\subsection{SENet}
Incorporating the SENet into weather forecasting models effectively intensifies their ability to handle and interpret the complexities of multi-channel datasets. The SENet architecture, with its global average pooling followed by a bottleneck of fully connected layers with Rectified Linear Unit (ReLU) and sigmoid activation function, allows the model to focus dynamically on the most relevant features across these channels. Since SENet's simplified architecture prevents the performance increase from the expense of excessive computing load, it is a useful addition to high-dimensional data processing in meteorological models \citep{iandola2016squeezenet}. By emphasizing significant features and suppressing lesser ones, SENet enhances model predictions of complex weather patterns owing to the nonlinear interactions among variables \citep{zhao2023spatiotemporal}.

Within the KARINA model, pointwise convolution is used for efficient channel integration and dimensionality refinement. The following integration of an SE layer effectively modifies the weight assigned to every channel, allowing the model to concentrate more on relevant channels. This method is quite similar to the variable aggregation strategy developed by Nguyen et al., where in cross-attention between channels, the approach minimizes processing demands while fostering a unified, context-aware representation at each spatial point \citep{nguyen2023climax}

\subsection{Geocyclic Padding}
GeoCyclic Padding is an innovative technique tailored for use with Convolutional Neural Networks (CNNs) on geospatial data, like the ERA5 dataset. This padding approach addresses the longitudinal continuity of the Earth by implementing circular padding along the horizontal edges of the input tensor, effectively connecting the 0° meridian with the 360° meridian. This mirroring of the Earth’s geography ensures a coherent representation of longitudinal direction, facilitating a continuous transition at the east-west edge. For the north-south edges, which correspond to the North and South poles, the method involves selecting a row from the circularly padded tensor, bisecting it at 180° longitude, and reordering the segments to reflect the data’s polarity. This step is crucial for accurately simulating the true continuity at the poles and minimizing the common distortions found in global map projections. By mirroring the Earth’s spherical geometry, GeoCyclic Padding allows for seamless transitions across both latitudinal and longitudinal boundaries, thus eliminating the artificial discontinuities of the traditional zero padding \citep{anderson2002maps}.

While Scher and Messori suggested Polar Padding, which was a pioneering solution specifically designed for handling the latitude-longitude grid of ERA5 by separately processing the Southern and Northern Hemispheres, thereby doubling the model's trainable parameters, GeoCyclic Padding presents a more resource-efficient method \citep{scher2024physics}. Unlike Polar Padding, GeoCyclic Padding applies a consistent padding technique across the dataset, ensuring spatial continuity without differentiating between hemispheres, presenting a more resource-efficient method. This approach avoids the significant parameter increase from the hemisphere-specific strategy, making GeoCyclic Padding an attractive option for ERA5 datasets.

\begin{figure}[h!]
	\centering
	\includegraphics[width=0.9\textwidth]{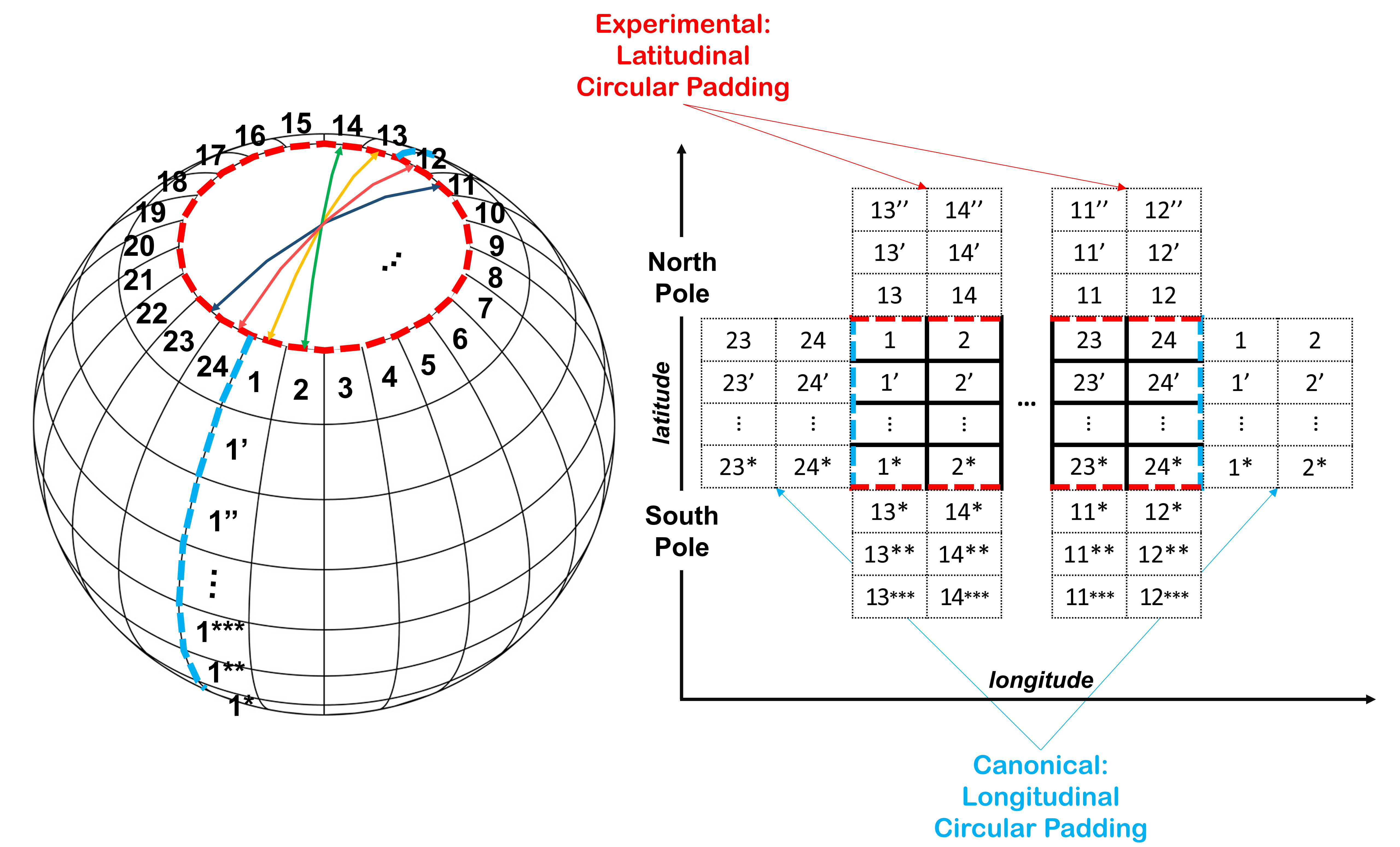}
	\caption{Description of GeoCyclic Padding. It minimizes projection distortions by circularly padding the edges and reordering at the poles to ensure longitudinal continuity in geographic data such as the ERA5 dataset.}
	\label{fig:fig2}
\end{figure}

\subsection{Training}
We used ERA5 data from 1979 to 2015 in the first training phase of our forecasting model, which was intended for one-day-ahead forecasts. Our approach was similar to that of FourCastNet \citep{pathak2022fourcastnet}. With the addition of orography data, this extensive dataset (66 variables) allows for an in-depth evaluation of the topographical effects on weather patterns. Employing the KARINA model, our approach analyzed daily datasets formatted as $X(k)$, where $k$ represents the time index with daily intervals. Specifically, the KARINA model was adeptly trained to predict the weather state one day ahead, $X(k+1)$, from the current day's state, $X(k)$, based on labeled data $X_{\text{label}}(k+1)$. The training procedure, conducted over 150 epochs with a learning rate of 0.001 and employing the AdamW optimizer in conjunction with a cosine learning rate adjustment, was efficiently completed in less than 12 hours with 4 NVIDIA A100 GPUs.

\subsection{Fine-Tuning}
Our model refinement strategy harnessed a specialized dataset enriched with time-lagged entries, extending from one to twenty-three hours, to enhance the temporal depth essential for forecasting precision \citep{cheon2024advancing}. Initially, our fine-tuning employed datasets at lag0 and lag12, doubling the data input, with a lower learning rate set at 0.005 to facilitate the model's adaptation to temporal data patterns \citep{nakamura2019revisiting}. Further refinement incorporated datasets at lag0, lag6, lag12, and lag18, augmenting the data volume fourfold, accompanied by a learning rate decrease to 0.0025. In the concluding phase, we integrated twenty-four time-lagged datasets, spanning the full hourly range up to lag23, and applied a reduced learning rate of 0.0001 to ensure meticulous calibration and tight alignment with the temporal intricacies inherent in the comprehensive lagged data spectrum.

\subsection{Evaluation Metrics}
In our study, we assessed the quality of our forecasts by employing latitude-weighted Root Mean Square Error (RMSE) and the Anomaly Correlation Coefficient (ACC). Our choice of these specific metrics is informed by their widespread use in previous related research, enabling us to conduct a fair and direct comparison with established studies. In equation (2), the upper bar denotes the mean and first three harmonics of the climatological seasonal cycle to exclude covariance of seasonality in calendar days. Then, the ACC assesses how well the forecast captures deviations from the expected climate.

% For RMSE
The Root Mean Square Error (RMSE) is defined as:
\begin{equation}
RMSE = \sqrt{\frac{1}{n} \sum_{i=1}^{n} (y_i - \hat{y}_i)^2}
\end{equation}

% For ACC
The formula for ACC is given by:
\begin{equation}
ACC = \frac{\sum_{i=1}^{n} (X_i - \bar{X})(Y_i - \bar{Y})}{\sqrt{\sum_{i=1}^{n} (X_i - \bar{X})^2 \sum_{i=1}^{n} (Y_i - \bar{Y})^2}}
\end{equation}

\section{Result}
\subsection{Effectiveness of Geocyclic Padding and Squeeze Net}
We conducted an in-depth investigation of KARINA variations used throughout seven-day predictions. We specifically concentrated on assessing the RMSE for key meteorological variables, such as MSL (mean sea level pressure), Z500 (geopotential at 500 hPa), T850 (air temperature at 850 hPa), and T2M (2 m air temperature). Three versions of the KARINA model were taken into consideration in our analysis: "KARINA Plain," "KARINA Padded," and "KARINA Padded+SENet." We saw prominent changes in forecasting accuracy between different configurations throughout the prediction period (Figure. 3). Among every configuration evaluated, the `KARINA Padded+SENet' variation consistently showed the lowest RMSE values (Table. 1). These results highlight the need to utilize advanced techniques to increase the dependability of weather forecasting models over long timelines by indicating that the addition of GeoCyclic Padding and SENet enhances prediction accuracy.

Table. 2 further investigated the influence of different kernel sizes in the Stem Layer. Building on existing studies that experimented with the FourCastNet model to enhance its performance, both our research and these studies have found that a smaller patch size leads to better outcomes. For instance, Cheon et al. identified a 1x1 patch size as optimal for the 2.5° resolution data. Our results supported this conclusion by demonstrating that the model performed substantially better with smaller kernels, as they focused on finer spatial details, especially when dealing with lower-resolution data \citep{cheon2024advancing,guo2024fourcastnext}

\begin{figure}[h!]
	\centering
	\includegraphics[width= 0.8\textwidth]{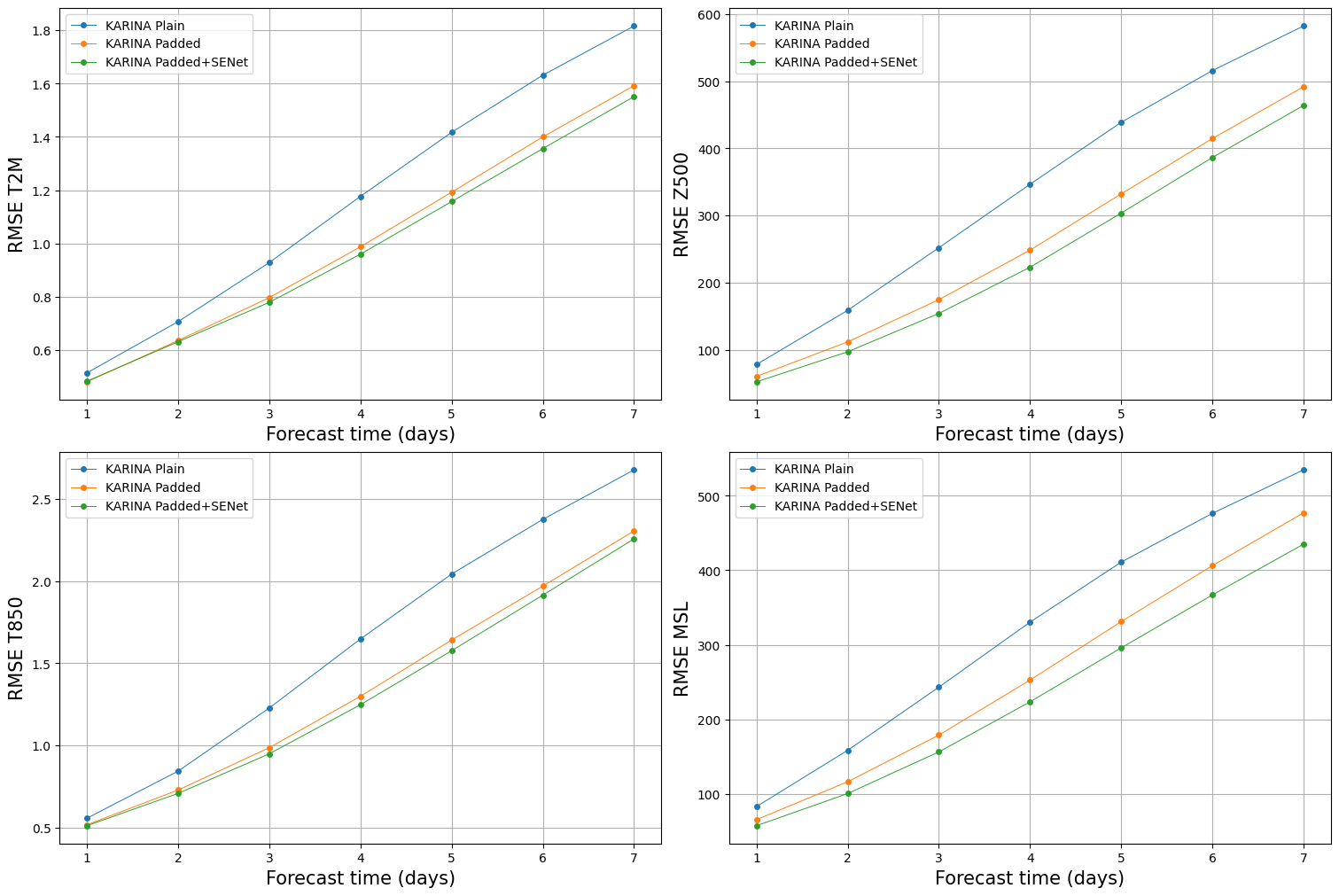}
	\caption{Comparative forecast skill evaluations of forecasting techniques over a 7-day horizon. The graphs display the globally averaged latitude-weighted RMSE for T2M, Z500, T850, and MSL over a 7-day forecast period. Three variations of the KARINA model are compared: `KARINA Plain' without any modifications, `KARINA Padded' incorporating GeoCyclic Padding, and `KARINA Padded+SENet' integrating both GeoCyclic Padding and SENet.}
	\label{fig:fig3}
\end{figure}

\begin{table}[h!]
  \caption{Impact of SENet and GeoCyclic Padding on Weather Forecasting Accuracy (RMSE) for Day 3 Predictions. The unit of each variable is indicated in the parenthesis.}
  \label{tab:weather-forecasting}
  \centering
  \begin{tabular}{@{}c@{\hspace{1em}}c@{\hspace{1em}}c@{\hspace{1em}}c@{\hspace{1em}}c@{\hspace{1em}}c@{}}
    \toprule
    SENet & GeoCyclic Padding & T2M (K) & MSL (Pa) & $Z500 (\text{m}^2 \text{s}^{-2})$ & T850 (K) \\
    \midrule
    - & - & 0.92 & 243.04 & 252.17 & 1.22 \\
    - & o & 0.79 & 178.74 & 174.92 & 0.98 \\
    o & o & 0.77 & 156.16 & 154.47 & 0.94 \\
    \bottomrule
  \end{tabular}
\end{table}

\begin{table}[h!]
  \caption{Same as Table 1, except for RMSE from KARINA+Padded+SENet with Varied Kernel Sizes in the Stem Layer.}
  \label{tab:kernel-size-impact}
  \centering
  \begin{tabular}{@{}c@{\hspace{1em}}c@{\hspace{1em}}c@{\hspace{1em}}c@{\hspace{1em}}c@{\hspace{1em}}c@{}}
    \toprule
    Kernel Size (Stem Layer) & T2M (K) & MSL (Pa) & $Z500 (\text{m}^2 \text{s}^{-2})$ & T850 (K) \\
    \midrule
    3 x 3 & 0.77 & 156.16 & 154.47 & 0.94 \\
    5 x 5 & 0.82 & 160.10 & 161.67 & 0.98 \\
    7 x 7 & 0.82 & 160.6 & 157.25 & 0.98 \\
    \bottomrule
  \end{tabular}
\end{table}
\FloatBarrier % Add this line right before your section to prevent floats from floating past it

The region-dependent response of the model variations was investigated in Figure. 4. The impact of GeoCyclic Padding yielded a pronounced improvement at the image boundary on forecast day 1, effective at both the meridian and the poles (Figure. 4a). This improvement spread as the forecast day increased, notably at the western boundary of mid-latitudes, where westerly atmospheric flow is crucial for prediction. SENet was more effective near the equator, where active atmospheric convection with vertical motion required improved channel integration (Figure. 4b). Figure. 4c describes ACC improvement from each model variation at different vertical levels. The ACC improvement from SENet was highest between the lower- and mid-troposphere (800–500 hPa), indicating better vertical level predictions of active tropical convection. The relatively smaller effect of SENet in the mid-latitude is presumably due to the dominant role of atmospheric horizontal flow rather than local convection. Overall, the model with SENet revealed reduced RMSE, where the GeoCyclic Padding impact was relatively small, indicating a uniform enhancement across regions compared to the plain model.
\begin{figure}[h!]
	\centering
	\includegraphics[width= 0.9\textwidth]{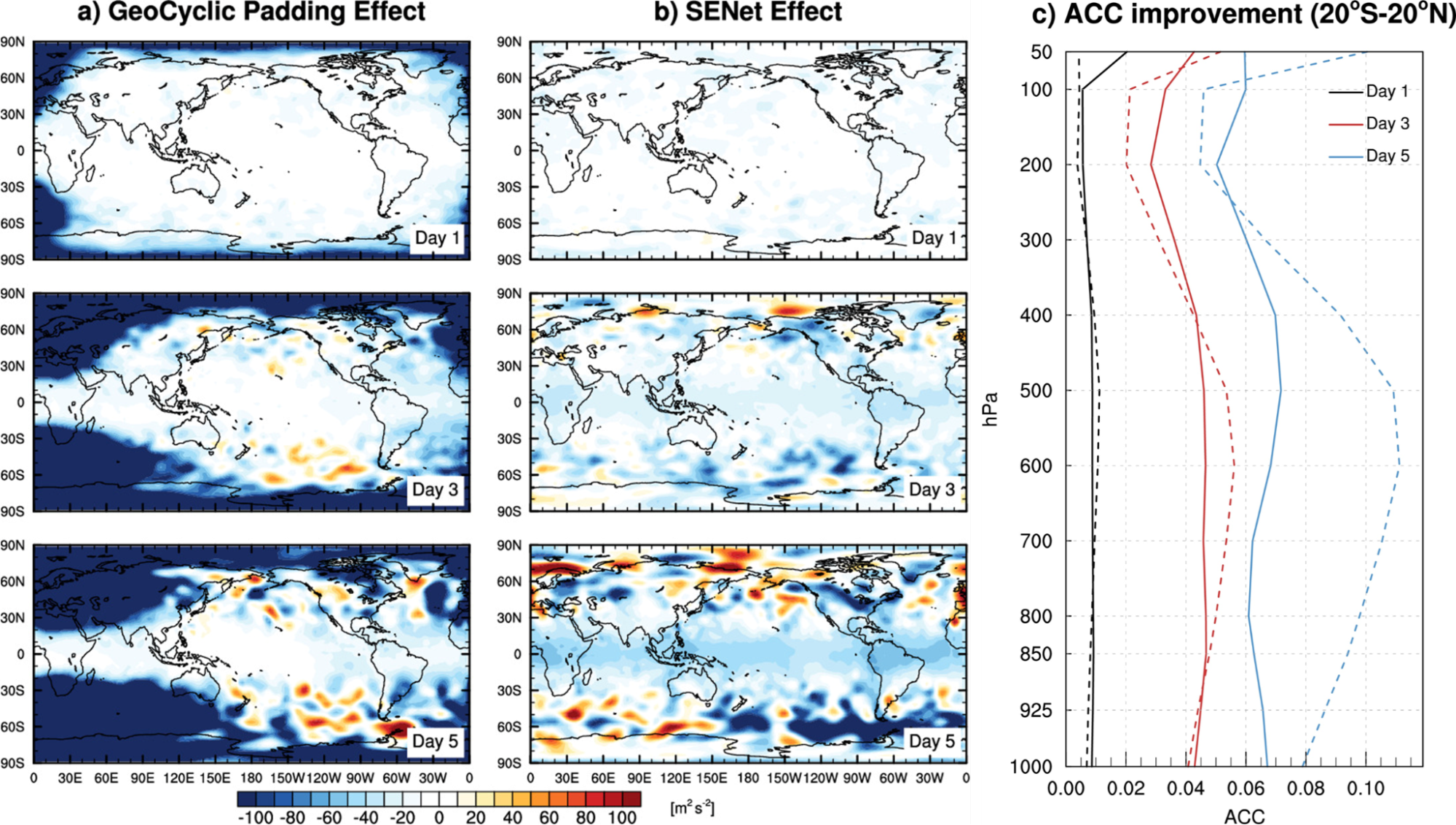}
	\caption{Horizontal distribution of prediction skill improvement during the test period. a) Effect of GeoCyclic Padding: RMSE difference between Padded and Plain. b) Effect of SENet: RMSE difference between Padded+SENet and Padded. c) ACC improvement of geopotential variable at each pressure level from GeoCyclic Padding (solid line) and SENet (dashed line) in the equatorial region (20°S–20°N).}
	\label{fig:fig4}
\end{figure}
\FloatBarrier % Add this line right before your section to prevent floats from floating past it

To see whether the relationship among the predicted variables has realistic atmospheric physics, a large-scale response from Z500 variation was investigated by linear regression among the test set prediction members as the following equation:
\begin{equation}
r_{zx} = \frac{\sum (Z'_{500NA} - \overline{Z'_{500NA}})(x' - \overline{x'})}{\sqrt{\sum (Z'_{500NA} - \overline{Z'_{500NA}})^2}}
\end{equation}
$Z'_{500\text{NA}}$ denotes a deviation of Z500 at forecast day 5 from the initial state, area-averaged in the North Atlantic region ($40^\circ\text{W} - 10^\circ\text{W};\ 30^\circ\text{N} - 45^\circ\text{N}$). The upper bar indicates an average of the prediction members, and $x'$ denotes the same deviation of the other variables for the regression. The regression coefficient $r_{zx}$ represents the systematic variation of $x'$ associated with the $z'_{500NA}$ during the 5-day prediction.
Fig 5 illustrates the regressed variation of the selected variables in the ERA5 and KARINA models, which indicates generalized large-scale atmospheric variation when a low-pressure system is dominant in the North Atlantic region. In ERA5, a high-pressure system near Norway and northerly flow brings cold temperatures in Northern Europe, which is a typical pattern known as the North Atlantic Oscillation \citep{kang2017increase}. The plain model reproduced the large-scale features as in ERA5, but the amplitude of the high-pressure system accompanied by cold anomalies in Northern Europe was much weaker than ERA5 (Fig 5b). With Geocyclic Padding, the amplitude variation in Northern Europe was well reproduced as ERA5 (Fig 5c). Additional improvements from SENet described more realistic features, including details such as the southwesterly flow and warming in Northern Africa (Fig 5d). This result demonstrated that the Geocyclic Padding and SENet components contribute to reproducing realistic large-scale atmospheric physics as well as improved skill metrics.

\begin{figure}[h!]
	\centering
	\includegraphics[width= 0.9\textwidth]{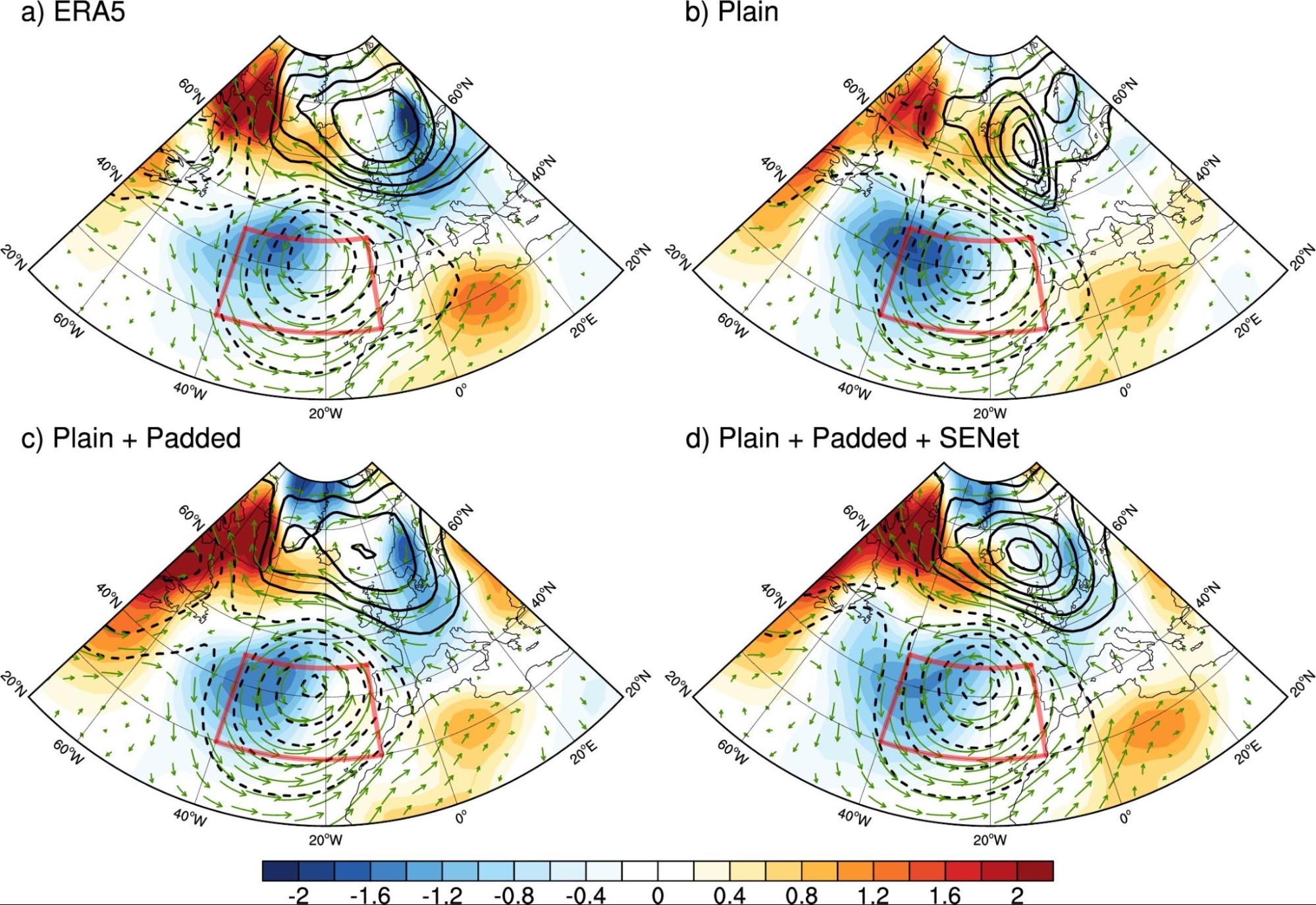}
	\caption{Regressed field of variables from $-\sigma$ of $z'_{500\text{NA}}$. Shaded colors represent the 2m temperature ($T2M$), contour lines represent mean sea level pressure ($MSL$), and green arrows indicate the 850 hPa wind vectors. The red boxes delineate the North Atlantic region ($40^\circ\text{W} - 10^\circ\text{W};\ 30^\circ\text{N} - 45^\circ\text{N}$).}
	\label{fig:fig4}
\end{figure}
\FloatBarrier % Add this line right before your section to prevent floats from floating past it

\subsection{Comparison with SOTA models}
The KARINA model's forecasting capabilities were compared with the ECMWF S2S model, which focuses on surface variables of TCWV and T2M rather than the pressure level variables unavailable for daily-mean. Fig 3 illustrates that KARINA accomplished significantly better in earlier prediction days due to the ECMWF S2S model's initial drift. KARINA maintained the lower forecast error of the three variables until day 7, indicating better prediction skills even after the initial condition effect diminished.  Furthermore, in comparative assessments to deep learning-based models, KARINA and its tuned one closely matched those of SOTA models including ClimaX, Pangu Weather, FourCastNet, and Graphcast as shown in Table 3. We noted that the other models calculated RMSE from the instantaneous data, apart from the daily-mean data used in ours. In this regard, we only compared the Z500 variable, which is less affected by intra-day variation that could largely affect the other variables, such as T2M \citep{seidel2005diurnal}. 

\begin{figure}[htbp]
	\centering
	\includegraphics[width=0.8\textwidth]{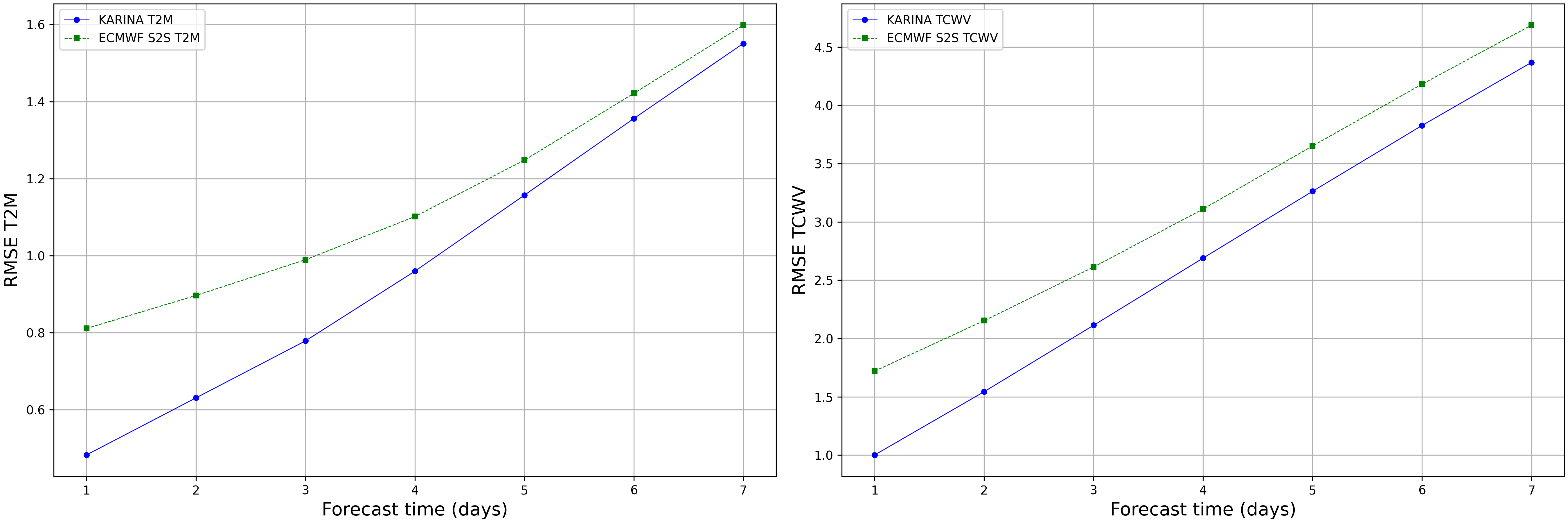}
	\caption{Comparative forecasting skill evaluations between KARINA and ECMWF S2S Models. The top row presents the RMSE across three meteorological variables over a forecast period of 7 days (TCWV, and T2M).}
	\label{fig:fig5}
\end{figure}

\begin{table}[htbp]
  \caption{Performance Comparison at deep learning-based SOTA models. The globally averaged RMSE for Z500 presented in this table was obtained from the study conducted by Nguyen et al \citep{nguyen2023climax}.}
  \label{tab:performance-comparison}
  \centering
  \begin{tabular}{@{}cccccccc@{}}
    \toprule
    Timestep & ClimaX & FCN(0.25) & Pangu & GraphCast & KARINA & KARINA\_Tuned \\
    \midrule
    24 hours (day 1)  & 96.19  & 81.31  & 42.23  & 38.77  & 52.78  & 52.12  \\
    72 hours (day 3) & 244.08 & 251.96 & 133.12 & 125.78 & 154.47 & 153.19 \\
    120 hours (day 5) & 440.4  & 483.44 & 295.63 & 271.65 & 303.64 & 301.75 \\
    168 hours (day 7) & 599.43 & 680    & 504.9  & 466.53 & 463.65 & 461.67 \\
    \bottomrule
  \end{tabular}
\end{table}
\FloatBarrier % Add this line right before your section to prevent floats from floating past it

\section{Conclusion}
We propose a data-driven model for global weather forecasting, named KARINA, that addresses the dual challenges of computational efficiency and forecasting accuracy. By integrating ConvNext, SENet, and Geocyclic Padding, it excelled at a 2.5° resolution and lead time of up to 7 days, outperforming the ECMWF S2S reforecasts and competing well with the recently developed deep learning-based high-resolution models. The Geocyclic Padding and SENet components of the model are vital to its achievement, providing notable enhancements in modeling Earth's atmosphere, specifically around the image edges and the equatorial regions, better representing the physics in the real world. With a better understanding of model components and physics representation, our efficient framework can accelerate the development of next-generation models for data-driven weather prediction. KARINA's success indicates a new era of efficiency and precision in climate research, offering a scalable and effective solution for advancing global weather forecasting.

\section{Acknowledgment}
This work was supported by the Ministry of Science and ICT through the National Research Foundation of Korea (NRF-2021R1C1C2004621 and NRF-2022M3K3A1094114). Computational resources have been supported by KISTI supercomputing center (Project No. KSC-2022-CRE-0521).
\clearpage 

\bibliographystyle{unsrtnat}
\bibliography{references}
\clearpage 

\section{Supplementary Materials}
\subsection{Detailed Experiment Setup}
\begin{table}[h!]
    \caption{List of ERA5 Atmospheric Variables used in KARINA, Corresponding Short Names, Vertical Pressure Levels in Hectopascals (hPa), and Standard Units of Measurement.}
    \centering
    \begin{tabularx}{\textwidth}{lXlX}
        \toprule
        \textbf{Variable name} & \textbf{Short name} & \textbf{Vertical levels (hPa)} & \textbf{Units} \\
        \midrule
        Zonal wind & U & 1000, 925, 850, 800, \\ & & 700, 600, 500, 400, \\ & & 300, 200, 100, 50 & m/s \\
        \addlinespace
        Meridional wind & V & 1000, 925, 850, 800, \\ & & 700, 600, 500, 400, \\ & & 300, 200, 100, 50 & m/s \\
        \addlinespace
        Temperature & T & 1000, 925, 850, 800, \\ & & 700, 600, 500, 400, \\ & & 300, 200, 100, 50 & K \\
        \addlinespace
        Specific humidity & Q & 1000, 925, 850, 800, \\ & & 700, 600, 500, 400, \\ & & 300, 200, 100, 50 & kg/kg \\
        \addlinespace
        Geopotential & Z & 1000, 925, 850, 800, \\ & & 700, 600, 500, 400, \\ & & 300, 200, 100, 50 & m\(^2\)/s\(^2\) \\
        \addlinespace
        2m temperature & T2m & \multicolumn{1}{c}{-} & K \\
        Mean sea level pressure & MSL & \multicolumn{1}{c}{-} & Pa \\
        Surface air pressure & SP & \multicolumn{1}{c}{-} & Pa \\
        \makecell[l]{Total column vertically-\\integrated water vapor} & TCWV & - & kg/m\(^2\) \\
        Skin temperature & SKT & - & K \\
        TOA incident solar radiation & TISR & \multicolumn{1}{c}{-} & J/m\(^2\) \\
        \bottomrule
    \end{tabularx}
\end{table}
\clearpage 

\begin{table}[htbp]
  \caption{Overview of Essential Hyperparameters in the KARINA Model for Global Weather Forecasting (Note: "dt" represents the prediction time interval of the trained model).}
  \label{tab:hyperparameters}
  \centering
  \begin{tabular}{lc}
    \toprule
    Hyperparameter & Value \\
    \midrule
    Loss & l2 \\
    Learning Rate (LR) & 0.001 \\
    Scheduler & CosineAnnealingLR \\
    dt & 1 day\\
    Number of In-Channels & 67 \\
    Number of Out-Channels & 67 \\
    Normalization & Z-score \\
    Optimizer Type & AdamW \\
    Max Epochs & 150 \\
    \bottomrule
  \end{tabular}
\end{table}
\FloatBarrier

\begin{table}[htbp]
  \caption{Comparative Analysis of Training Times and Hardware Specifications for Deep Learning Models.}
  \label{tab:training-details}
  \centering
  \begin{tabular}{lcc}
    \toprule
    Model & Number of GPUs & Training Time \\
    \midrule
    FuXi \citep{chen2023fuxi} & 8 Nvidia A100 GPUs & 30 hrs \\
    Fengwu \citep{chen2023fengwu} & 32 Nvidia A100 GPUs & 17 days \\
    FourCastNet \citep{pathak2022fourcastnet} & 64 Nvidia A100 GPUs & 16 hrs \\
    GraphCast \citep{lam2022graphcast}& 32 Google Cloud TPU v4 devices & 4 weeks \\
    Pangu Weather \citep{bi2023accurate}& 192 NVIDIA Tesla-V100 GPUs & 64 days \\
    KARINA & 4 Nvidia A100 GPUs & 12 hrs \\
    \bottomrule
  \end{tabular}
\end{table}
\FloatBarrier

\subsection{Comparison with SOTA models}
Tables 7, 8, and 9 present a comprehensive summary of the global forecasting performance of the KARINA model across some target variables, including T2M, Z500, and T850 at various lead times. These results were compared against an array of state-of-the-art models, namely ClimaX, FourCastNet, Pangu Weather, Graphcast, and HRES \citep{nguyen2023climax, lam2022graphcast, bi2023accurate, pathak2022fourcastnet}. This specificity facilitated a more straightforward comparison with established benchmarks, thereby enhancing the clarity and relevance of the findings within the context of global weather forecasting advancements. Despite being trained on a significantly lower resolution than its counterparts, except for ClimaX, KARINA achieved performance that was not only comparable but in some cases superior to the other models across different lead times. This outcome indicated the effectiveness and efficiency of the KARINA model. It is crucial, however, to acknowledge that while the other models derived their RMSE from instantaneous data, our analysis utilized daily-mean data.

\begin{table}[htbp]
  \caption{Performance Comparison of deep learning-based SOTA models at various spatial resolutions. The globally averaged RMSE for T2M was presented, with ClimaX at 5.625°, FourCastNet at 0.25°, Pangu at 0.25°, GraphCast at 0.25°, HRES at 0.1°, KARINA and KARINA\_Tuned at 2.5° resolution. Data was obtained from the study conducted by Nguyen et al \citep{nguyen2023climax}.}
  \label{tab:performance-comparison}
  \centering
  \begin{tabular}{@{}cccccccc@{}}
    \toprule
    Timestep & ClimaX & FCN & Pangu & GraphCast & HRES & KARINA & KARINA\_Tuned\\
    \midrule
    24 hours (day 1) & 1.1 & 0.95 & 0.72 & 0.62 & 0.66 & 0.48 & 0.48 \\
    72 hours (day 3) & 1.43 & 1.38 & 1.05 & 0.95 & 1.06 & 0.77 & 0.77 \\
    120 hours (day 5) & 1.83 & 1.99 & 1.53 & 1.36 & 1.52 & 1.15 & 1.14 \\
    168 hours (day 7) & 2.18 & 2.54 & 2.06 & 1.88 & 2.06 & 1.55 & 1.54 \\
    \bottomrule
  \end{tabular}
\end{table}
\FloatBarrier % Add this line right before your section to prevent floats from floating past it

\begin{table}[htbp]
  \caption{RMSE for Z500 from the same models and settings as Table 7.}
  \label{tab:performance-comparison}
  \centering
  \begin{tabular}{@{}cccccccc@{}}
    \toprule
    Timestep & ClimaX & FCN & Pangu & GraphCast & HRES & KARINA & KARINA\_Tuned\\
    \midrule
    24 hours (day 1) & 96.19 & 81.31 & 42.23 & 38.77 & 45.9 & 52.78 & 52.12 \\
    72 hours (day 3) & 244.08 & 251.96 & 133.12 & 125.78 & 146.37 & 154.47 & 153.19 \\
    120 hours (day 5) & 440.4 & 483.44 & 295.63 & 271.65 & 316.79 & 303.64 & 301.75 \\
    168 hours (day 7) & 599.43 & 680 & 504.9 & 466.53 & 553.93 & 463.65 & 461.67 \\
    \bottomrule
  \end{tabular}
\end{table}
\FloatBarrier

\begin{table}[htbp]
  \caption{RMSE for T850 from the same models and settings as Table 7.}
  \label{tab:performance-comparison}
  \centering
  \begin{tabular}{@{}cccccccc@{}}
    \toprule
    Timestep & ClimaX & FCN & Pangu & GraphCast & HRES & KARINA & KARINA\_Tuned\\
    \midrule
    24 hours (day 1) & 1.11 & 0.81 & 0.72 & 0.58 & 0.7 & 0.51 & 0.51 \\
    72 hours (day 3) & 1.59 & 1.55 & 1.13 & 1.02 & 1.27 & 0.94 & 0.94 \\
    120 hours (day 5) & 2.23 & 2.47 & 1.78 & 1.63 & 1.96 & 1.57 & 1.56 \\
    168 hours (day 7) & 2.77 & 3.3 & 2.6 & 2.41 & 2.78 & 2.25 & 2.24 \\
    \bottomrule
  \end{tabular}
\end{table}
\FloatBarrier

\subsection{GeoCyclic Padding Effect at the Poles}
In this study, we introduced GeoCyclic Padding, which covers both the latitudinal and longitudinal edges of the image, significantly enhancing global weather prediction accuracy. To examine its impact, we compared GeoCyclic Padding with `Circular Padding,' a technique traditionally used in Earth models. Notably, Circular Padding applies zero padding at both the southern and northern edges. To achieve this, we specifically isolated the effect of GeoCyclic Padding at the poles, enabling a focused assessment of its performance within the polar region. The partial effect of Geocyclic Padding at the latitudinal edge was evaluated by a difference of RMSE between Geocyclic Padding and Circular Padding in Figure 7. The partial Geocyclic Padding effect revealed a noticeable error reduction near the pole on forecast day 1. On day 3, this improvement spread to mid-latitude regions during autoregressive inference, achieving heightened accuracy. This result demonstrated the effective contribution of latitudinal edges in Geocyclic Padding, distinguished from Circular Padding.  
\begin{figure}[h!]
	\centering
	\includegraphics[width= 0.9\textwidth]{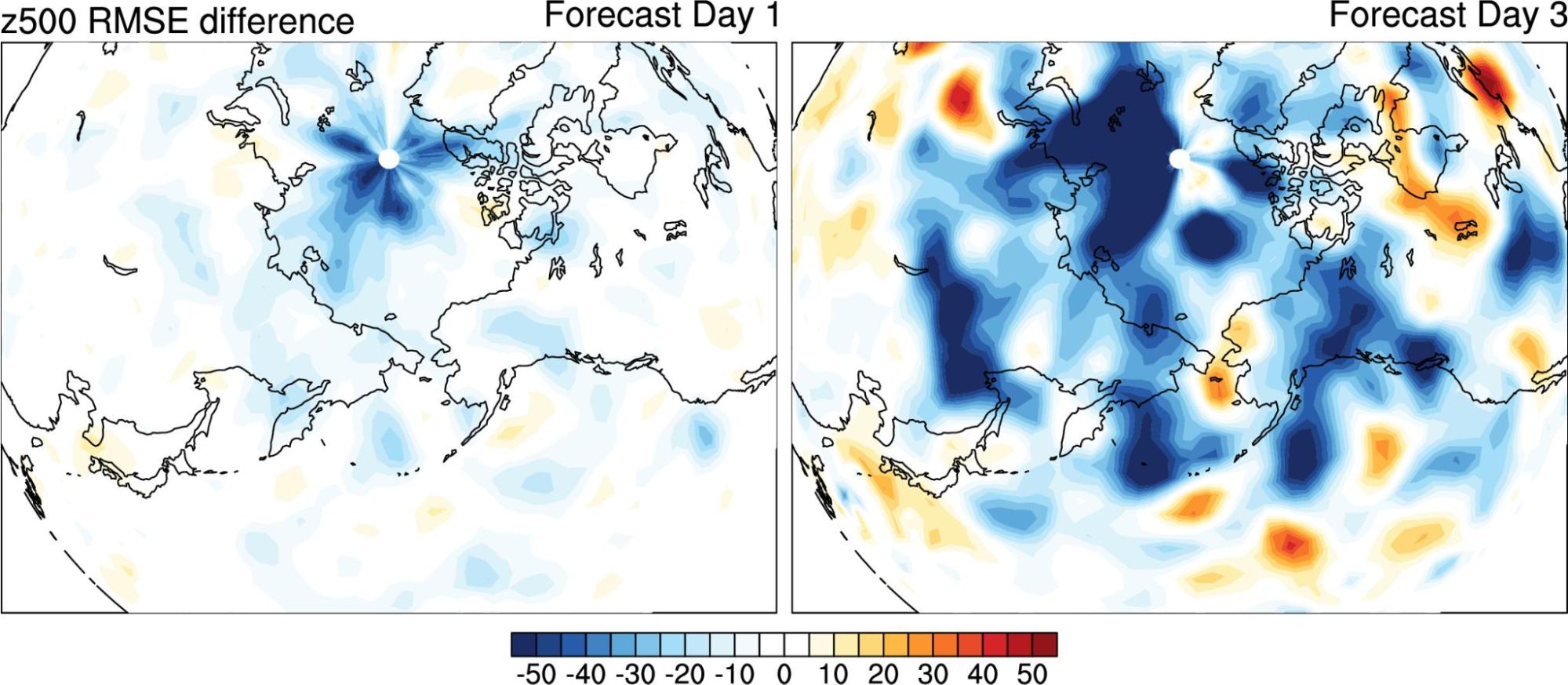}
	\caption{Horizontal distribution of RMSE improvement contributed by the latitudinal edge of Geocyclic Padding, which is defined as the RMSE difference of z500 between Geocyclic Padding and Circular Padding during the test period.}
	\label{fig:fig4}
\end{figure}
\FloatBarrier % Add this line right before your section to prevent floats from floating past it

\subsection{Stable autoregressive inference}
Occasionally, a repetitive autoregressive inference of global weather prediction reveals instability in some models, which could diminish the efficacy of data-driven models for reliable long-range forecasts. Several models have exhibited stable inference capabilities extending up to a year \citep{weyn2021sub,bonev2023spherical}, while most models have not been able to maintain consistent stability. Figure 8 exhibits the reproduced TCWV variable, total water vapor amount in the atmospheric column, captured by the autoregressive inference of KARINA beginning on Jan. 1st, 2018. Compared with ERA5 as ground truth, long-range prediction of KARINA revealed a realistic distribution of global atmospheric water vapor. Although detailed patterns on days 90 and 180 differed from ERA5 due to the less predictability in these forecast days, KARINA successfully reproduced a stable global pattern over six months.		 			 		

\begin{figure}[h!]
	\centering
	\includegraphics[width= 0.9\textwidth]{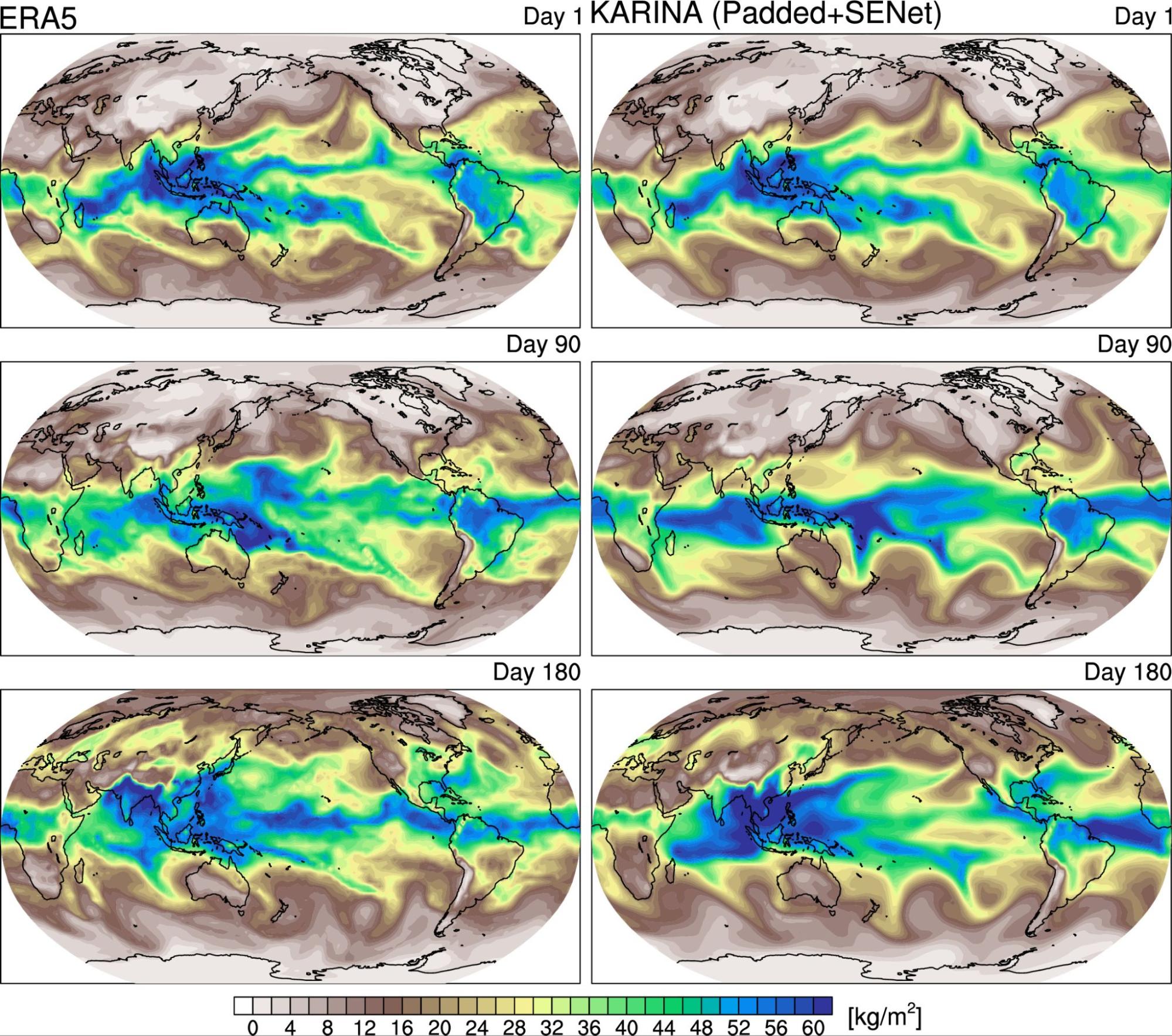}
	\caption{Horizontal distribution of the TCWV variable, as obtained from KARINA's autoregressive inference at various temporal milestones (day 1, day 90, and day 180), is collocated with ERA5 data for corresponding dates starting from January 1st, 2018.}
	\label{fig:fig4}
\end{figure}
\FloatBarrier % Add this line right before your section to prevent floats from floating past it

\clearpage
\subsection{A coherent proficient prediction of all variables}
Figure 9 illustrates the ACC skill scores for the various predictive variables within our model, extending through a 10-day forecast period. The proposed model demonstrated competent forecasting capabilities across all variables within the first week. Specifically, it exhibited enhanced proficiency for pressure-level variables, with particularly adept forecasting observed in the upper troposphere compared to surface levels. This result indicated that additional effort is necessary to improve the surface boundary condition in the upcoming models.

\begin{figure}[htbp]
	\centering
	\includegraphics[width=0.45\textwidth, height=17cm]{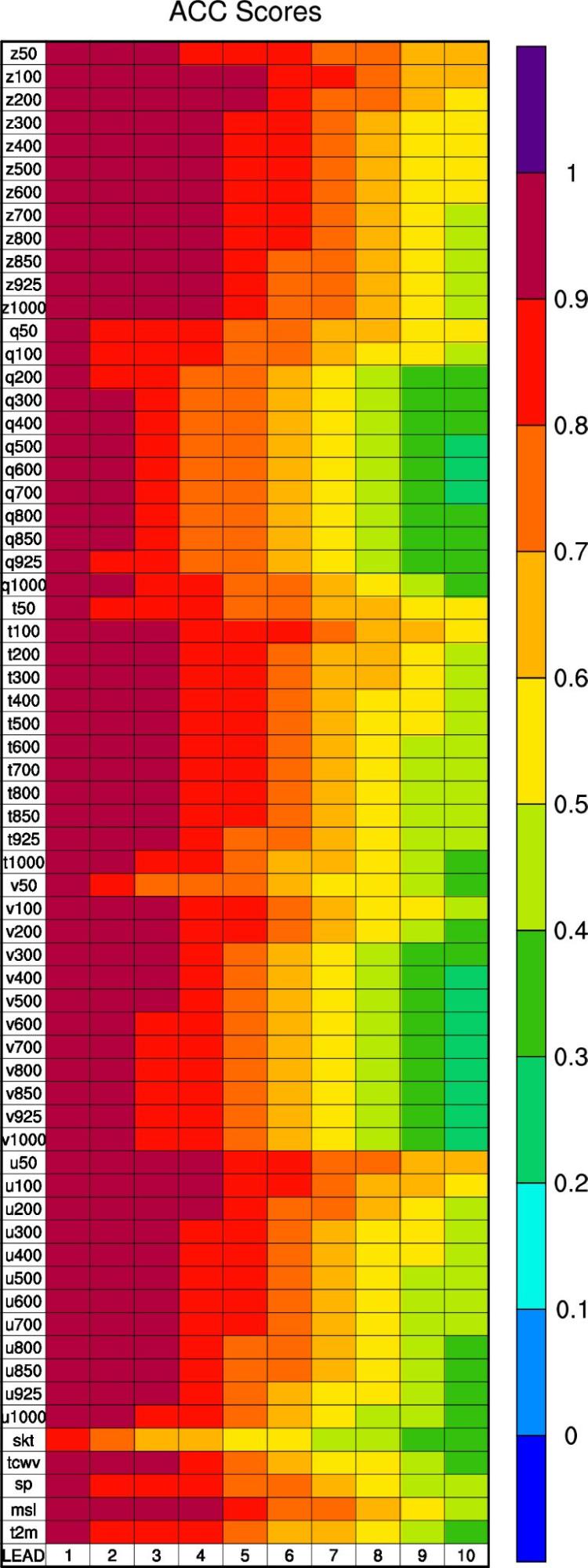}
	\caption{ACC skill scores for forecast variables up to day 10 (KARINA tuned).}
	\label{fig:fig5}
\end{figure}

\end{document}